\documentclass{article}
\usepackage{ijcai24}

\usepackage{times}
\usepackage{soul}
\usepackage{url}
\usepackage[hidelinks]{hyperref}
\usepackage[utf8]{inputenc}
\usepackage[small]{caption}
\usepackage{graphicx}
\usepackage{amsmath}
\usepackage{amsthm}
\usepackage{booktabs}
\usepackage{algorithm}
\usepackage{algorithmic}
\usepackage[switch]{lineno}

\usepackage[dvipsnames]{xcolor}
\usepackage{xspace}

\definecolor{SHINSYU}{HTML}{AB3B3A}

\definecolor{F7E0D5}{RGB}{247,224,213}
\definecolor{darkF7E0D5}{RGB}{209,154,128}
\colorlet{Light}{White!0!F7E0D5}
\usepackage{pifont}
\newcommand{\xmark}{\ding{55}}

\newcommand{\rownumber}[1]{\textcolor{darkF7E0D5}{#1}}

\newcommand{\setting}{PTTA\xspace}
\newcommand{\method}{ResiTTA\xspace}

\newcommand{\VspaceBefore}{\vspace{-3mm}}
\newcommand{\VspaceAfter}{\vspace{-3mm}}

\newcommand{\argmax}{\mathop{\arg\max}}

\usepackage{amssymb}
\usepackage{amsmath}
\usepackage{amsthm}
\usepackage{bbm}

\theoremstyle{definition}

\usepackage{float}

\usepackage{bibunits}
\nolinenumbers

\title{Resilient Practical Test-Time Adaptation: Soft Batch Normalization Alignment and Entropy-driven Memory Bank}

\author{
Xingzhi Zhou$^{1,3}$\thanks{This work was done while Xingzhi was an intern at
NVIDIA}
\and
Zhiliang Tian$^2$\thanks{Corresponding Author} \and
Ka Chun Cheung$^{3}$\and
Simon See$^3$\and
Nevin L. Zhang$^1$\\
\affiliations
$^1$Department of Computer Science and Engineering, HKUST\\
$^2$ College of Computer, NUDT\\
$^3$NVIDIA AI Technology Center, NVIDIA\\
\emails
xingzhi.zhou@connect.ust.hk,
tianzhiliang@nudt.edu.cn,
\{chcheung,ssee\}@nvidia.com,
lzhang@cse.ust.hk
}

\begin{document}

\maketitle

\begin{abstract}
    Test-time domain adaptation effectively adjusts the source domain model to accommodate unseen domain shifts in a target domain during inference. However, the model performance can be significantly impaired by continuous distribution changes in the target domain and non-independent and identically distributed (non-i.i.d.) test samples often encountered in practical scenarios. While existing memory bank methodologies use memory to store samples and mitigate non-i.i.d. effects, they do not inherently prevent potential model degradation. To address this issue, we propose a resilient practical test-time adaptation (\method) method focused on parameter resilience and data quality. Specifically, we develop a resilient batch normalization with estimation on normalization statistics and soft alignments to mitigate overfitting and model degradation. We use an entropy-driven memory bank that accounts for timeliness, the persistence of over-confident samples, and sample uncertainty for high-quality data in adaptation.  Our framework periodically adapts the source domain model using a teacher-student model through a self-training loss on the memory samples, incorporating soft alignment losses on batch normalization. We empirically validate \method across various benchmark datasets, demonstrating state-of-the-art performance. 

\end{abstract}

\section{Introduction}   
Test-time domain adaptation (TTA) updates a source-pretrained model to a target domain during the inference stage by adjusting parameters using only unlabeled test data streams. Due to the domain shift between source training data and target test data, domain adaptation is essential for achieving superior performance. This makes TTA crucial for the practical deployment of machine perception applications confronting domain shifts. For instance, a semantic segmentation model trained on a dataset in clear weather conditions performs poorly in rainy conditions \cite{hamlet_iccv_23}. Similarly, a pre-trained image classification model might experience degraded performance when tested on corrupted images generated from sensor-degraded cameras. 

Continuous distribution changes in the target domain and temporal correlation present significant challenges in practical model deployment, particularly for traditional TTA methods that assume test samples are sampled from a fixed target domain distribution. These continuous distribution changes require a TTA method to adjust parameters dynamically and prevent catastrophic forgetting. The temporal correlation contributes to non-independent and identically distributed (non-i.i.d.) test samples, where test samples from certain classes may predominate in specific time slots. 

Early TTA approaches have employed entropy minimization \cite{tent_iclr_21} and pseudo-labels \cite{pseudo_labeling} to adjust the model to target domains. However, these strategies might overfit to particular domains, resulting in model performance decline when the distribution continuously changes in the target domain. CoTTA \cite{cotta_cvpr_22} addresses continuous distribution changes. It utilizes a teacher-student model combined with augmentation-average methods to generate pseudo labels and stochastic parameter restoration to reduce overfitting. Nonetheless, CoTTA faces challenges with non-independent and identically distributed (non-i.i.d.) test samples during inference, as it is designed under the assumption of i.i.d. \cite{note_neurips_22,rotta_cvpr_23}. NOTE \cite{note_neurips_22} introduces a memory bank strategy to handle the non-i.i.d. effect observed in inference with instance-aware batch normalization. This method suggests a balance between individual instance statistics and source statistics within batch normalization. However, instance statistics can be quite unstable, resulting in model degradation. RoTTA \cite{rotta_cvpr_23} enhances the memory bank approach by accounting for sample timeliness and uncertainty, and it stabilizes batch normalization statistics by continually updating global statistics with test-batch statistics. Still, existing memory-bank techniques do not fully address potential model degradation caused by overfitting on test samples in the target domain.

In this paper, to tackle potential model degradation, we propose a resilient practical test-time adaptation (\method) method. This method consists of three parts: resilient batch normalization, entropy-driven memory bank, and self-training adaptation. Resilient Batch Normalization (ResiBN), involves maintaining global target statistics for batch normalization. We update these statistics using an exponential moving average based on test-batch statistics. To prevent parameter overfitting, ResiBN employs soft alignments on the target statistics by minimizing the Wasserstein distance between the target statistics and the source statistics, where the source statistics are global statistics in batch normalization acquired during the source model training. Entropy-driven Memory Bank (EntroBank), updates samples by considering three factors: timeliness,  the persistence of over-confident samples, and sample uncertainty to ensure high data quality. Timeliness addresses the issue of outdated samples that remained in the memory for too long. The persistence of over-confident samples refers to low-entropy samples over an extended period. Sample uncertainty is measured using the entropy of the predicted distribution as a metric for memory updates. In the adaptation stage, we periodically adapt the source pre-trained model using a teacher-student approach with samples from EntroBank via a self-training loss, coupled with soft alignment losses in batch normalization statistics. 
Our contributions are as follows: 
\begin{itemize}
    \item  We propose a resilient practical test-time adaptation approach (\method) to counter potential model degradation under continuous distribution changes in the target domain and temporal correlation. 
    \item To reduce parameter overfitting, we propose Resilient Batch Normalization which includes gradually updated target statistics in batch normalization and employs soft alignments on these statistics. 
    \item To enhance data quality, we propose Entropy-Driven Memory Bank that considers the timeliness of data, the persistence of over-confident samples, and sample uncertainty. 
    \item  We conduct comprehensive experiments to validate \method on several common TTA benchmarks such as CIFAR10-C, CIFAR100-C, and ImageNet-C, where \method outperforms existing state-of-the-art results.
\end{itemize}

\section{Related Works}

\subsection{Domain Adaptation}
Domain adaptation (DA) concentrates on transferring knowledge from a source domain to a target domain \cite{Tzeng_2015_ICCV,JMLR:v17:15-239,Tsai_2018_CVPR,Li_2021_CVPR}. It falls into two categories: supervised and unsupervised, determined by the labeling status of the target domain data. The key approaches in DA includes latent distribution alignments \cite{8454781,Xu_2019_ICCV}, adversarial training \cite{JMLR:v17:15-239,Saito_2018_CVPR}, self-training \cite{Zou_2018_ECCV,10018569}, etc. Traditional DA techniques require access to both labeled source datasets and target datasets during training, which may limit their practical use when target domain data is unavailable. This limitation has spurred interest in test-time domain adaptation. 

\subsection{Test-Time Domain Adaptation}
Test-time domain adaptation (TTA) aims to improve model performance at test time by adapting the model to the target domain distribution, utilizing only the source model and unlabeled target data \cite{Chi2021TestTimeFast,TTT_nips_21,ijcai2021p402,Chen2022ContrastiveTestTime,Jang2023TestTimeAdaptation,SAR_ICLR_2023}. TTA is categorized into offline and online settings. The offline setting allows the model to access all test samples, whereas the online setting processes test data in batches, more practical for real-world deployment. \cite{Schneider2020Improvingrobustnesscommon} demonstrates that building batch normalization statistics at test time significantly improves performance under domain shifts. TENT \cite{tent_iclr_21} illustrates that adjusting batch normalization parameters with entropy minimization is effective for single target domain adaptation. EATA \cite{Niu2022EfficientTestTime} expands on this concept by incorporating weighted entropy minimization that accounts for reliability and diversity. It further applies elastic weight consolidation \cite{Kirkpatrick2017Overcomingcatastrophicforgetting} from the continual learning field to mitigate catastrophic forgetting in the source domain.

Traditional TTA assumes test data is sampled from a fixed target domain. However, the target domain often experiences continuous distribution shifts in real-world scenarios. CoTTA \cite{cotta_cvpr_22} addresses this challenge in TTA, termed continual test-time adaptation (CTTA). It employs a teacher-student model with augmentation-average pseudo labels and stochastic weight recovery to reduce overfitting. RMT \cite{Doebler2023Robustmeanteacher} employs symmetric cross-entropy and contrastive learning for robust training. LAME \cite{lame_cvpr_22}, NOTE \cite{note_neurips_22}, and RoTTA \cite{rotta_cvpr_23} tackle temporal correlation in TTA. LAME \cite{lame_cvpr_22} fixes the source pre-trained model and modifies prediction probabilities by imposing a Laplacian constraint on the test batch. NOTE \cite{note_neurips_22} employs a memory bank to manage non-i.i.d. test samples and introduces instance-aware batch normalization for adaptation. RoTTA \cite{rotta_cvpr_23} handles both continuous distribution changes and temporal correlation in the target domain, forming the new problem setup practical test-time adaptation (PTTA). It uses a memory bank considering timeliness, category balance, and sample uncertainty. It also maintains global statistics in batch normalization continually updated with test-batch statistics. Nevertheless, existing memory bank methods may not fully address model degradation under simultaneous temporal correlation and continuous distribution changes. In this work, we introduce \method to tackle potential model degradation.

\section{Method}

\subsection{Problem Definition}
We adhere to the problem setup in practical test time adaption (PTTA) \cite{rotta_cvpr_23}. Given a source pre-trained model $f_{\theta_s}$ with parameter $\theta_s$ pre-trained on the source domain dataset $\mathcal{\mathcal{D}}_s=\{(x_s,y_s)\}$, the objective is to adapt $f_{\theta_s}$ to a sequence of online unlabeled samples $\mathcal{X}_1, \mathcal{X}_2,...,\mathcal{X}_T$ . Each $\mathcal{X}_t$ represents a batch of temporally correlated samples from a continuously evolving distribution $\mathcal{P}_t$. The goal is to make inference $f_{\theta_t}(\mathcal{X}_t)$ using the adapted parameter $\theta_t$ at each time $t$.

\subsection{Resilient Practical Test Time Adaptation}
\begin{figure*}
    \centering
    \VspaceBefore
    \includegraphics[width=0.9\linewidth]{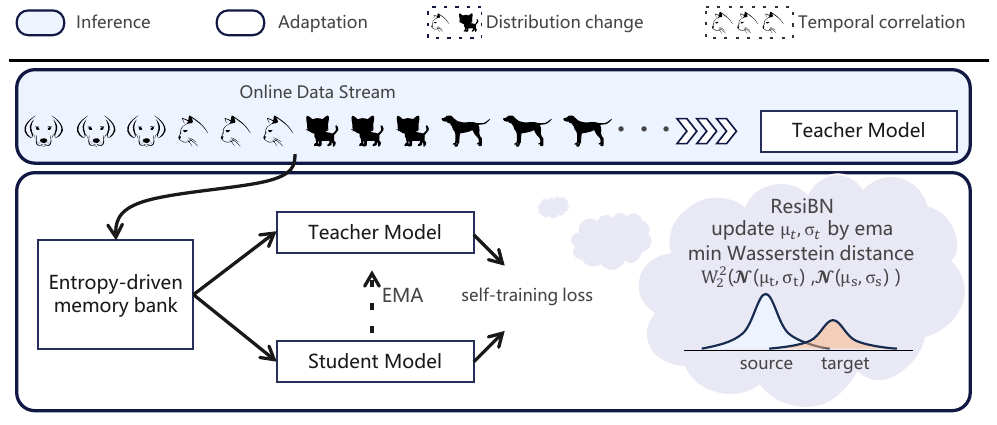}
    \VspaceAfter
    \caption{\textbf{Framework Overview.} We duplicate the pre-trained model into a student and a teacher model at the beginning of the test and replace the batch normalization layer with resilient batch normalization (ResiBN). In the inference stage, we use the teacher model to predict labels. In the adaptation stage, we collect online test streams by entropy-drive memory bank (EntroBank). We periodically adapt the model through a self-training loss on data drawn from the memory bank, incorporating soft alignment losses.}
    \label{fig:diagram}
\end{figure*}

Motivated by the need to protect current models against potential deterioration due to continuous distribution changes in the target domain and temporal correlation, we develop resilient batch normalization (ResiBN) by gradually updating global target statistics and employing soft alignments by minimizing the Wasserstein distance between target statistics and source statistics. To get high-quality data for effective adaptation, we introduce an entropy-driven memory bank (EntroBank) factoring in timeliness, the persistence of over-confident samples, and sample uncertainty. In the adaptation stage, we employ a teacher-student method to periodically adapt the model to the target domain using memory samples through a self-training loss, complemented with soft alignment losses on the target statistics. We discuss ResiBN in sec. \ref{sec:resilient_BN}, EntroBank in sec. \ref{sec:EntroBank} and adaptation using a teacher-student model in sec. \ref{sec:self-training}. The framework overview of \method is presented in Fig. \ref{fig:diagram}. 

\subsection{Resilient Batch Normalization }
\label{sec:resilient_BN}
Resilient Batch Normalization (ResiBN) maintains global target statistics $\mu_t, \sigma_t^2$, which are gradually updated by the test batch statistics in the batch normalization. Concurrently, ResiBN employs soft alignments on these target statistics by minimizing the Wasserstein distance between $\mathcal{N}(\mu_t, \sigma_t^2), \mathcal{N}(\mu_s, \sigma_s^2)$, where $(\mu_s,\sigma_s^2)$ denote the source statistics acquired in batch normalization during the source training stage. 

In a 2-D batch normalization (BN) scenario with a given feature map $\boldsymbol{X}\in\mathbb{R}^{B \times C \times H \times W}$, the BN calculates mean $\mu \in \mathbb{R}^C$ and variance $\sigma^2 \in \mathbb{R}^C$ across B, H, W axes.  

\begin{align}
\mu_c & =\frac{1}{B H W} \sum_{b=1}^B \sum_{h=1}^H \sum_{w=1}^W \boldsymbol{X}_{(b, c, h, w)}, \\
\sigma^2_c & =\frac{1}{B H W}\sum_{b=1}^B \sum_{h=1}^H \sum_{w=1}^W\left(\boldsymbol{X}_{(b, c, h, w)}-\mu_c\right)^2 .
\end{align}

Subsequently, the feature map undergoes channel-wise normalization:  
\begin{equation}
B N\left(\boldsymbol{X}_{(b, c, h, w)} ; \mu, \sigma^2\right)=\gamma_c \frac{\boldsymbol{X}_{(b, c, h, w)}-\mu_c}{\sqrt{\sigma^2_c+\epsilon}}+\beta_c,
\end{equation}
where $\gamma, \beta \in \mathbb{R}^C$ are learnable affine parameters post-normalization, enhancing learning ability in the BN layer. $\epsilon$ is a small constant for numerical stability. In training, BN maintains global running mean and variance values $\left(\mu_s, \sigma_s^2\right)$, termed source statistics in BN, measured by the exponential moving average of training batch statistics.

ResiBN maintains global target statistics $\mu_t,\sigma_t$ in batch normalization to stably estimate statistics amidst continual distribution changes and temporal correlation, following \cite{rotta_cvpr_23}, $\mu_t, \sigma_t$ are updated by the test batch statistics:

\begin{align}
\mu_t & =(1-\nu_b) \mu_t+\nu_b \mu_b, \\
\sigma_t^2 & =(1-\nu_b) \sigma_t^2+\nu_b \sigma_b^2,
\end{align}

where $\mu_b,\sigma_b^2$ are the test batch statistics. $\nu_b$ is the parameter controlling the update rate. 

Continuous distribution changes in the target domain might lead to overfitting on the target statistics $\mu_t, \sigma_t$. To enhance parameter resilience, we introduce a regularization of the target statistics inspired by continual learning \cite{Kirkpatrick2017Overcomingcatastrophicforgetting}. We consider various divergence measures between source and target distribution, such as KL and JS divergence. However, we choose to minimize the Wasserstein distance \cite{villani2009optimal} for numerical stability in BN. The reasoning for this choice over KL and JS divergence for soft alignments is elaborated in the appendix.

Soft alignments on the target statistics are achieved by minimizing Wasserstein distance:
\begin{align}
    \min W_2^2(\mathcal{N}(\mu_t,& \sigma_t^2), \mathcal{N}(\mu_s, \sigma_s^2))  \nonumber \\ 
    &=\left(\mu_s-\mu_t\right)^2 + \sigma_s^2+\sigma_t^2-2 \sigma_s\sigma_t.
\end{align}

The derivative with respect to $\mu_t$ and $\sigma_t^2$ are:
\begin{equation}
    \frac{dW_2^2}{d\mu_t} = 2(\mu_t -\mu_s),
\end{equation}
\begin{equation}
\label{eq:derivative_sigma_squre}
     \frac{dW_2^2}{d\sigma_t^2}  = 1-\frac{\sigma_s}{\sigma_t}.
\end{equation}

Given that $\sigma_t$ appears in the denominator in Eq. \ref{eq:derivative_sigma_squre}, which could cause numerical instability. We use the derivative with respect to $\sigma_t$:
\begin{equation}
    \frac{dW_2^2}{d\sigma_t} = 2\sigma_t-2\sigma_s.
\end{equation}

Finally,  $\mu_t, \sigma_t$ are updated for soft alignment with the source statistics as follows:
\begin{equation}
    \mu_t = \mu_t -\eta_t \frac{dW_2^2}{d\mu_t} ,
\end{equation}
\begin{equation}
    \sigma_t = \sigma_t - \eta_t \frac{dW_2^2}{d\sigma_t} ,
\end{equation}

where $\eta_t$ denotes hyperparameters indicating the extent of regularization.

\subsection{Entropy-driven Memory Bank}
\label{sec:EntroBank}

Entropy-driven Memory Bank (EntroBank) focuses on three aspects: timeliness, the persistence of over-confident samples, and sample uncertainty. Due to continual distribution changes, our priority is to remove outdated samples for timeliness. We then prioritize the removal of long-persisted over-confident samples to avoid overfitting. In the absence of outdated or long-persisted over-confident samples, we determine sample removal based on a comparison of sample uncertainty measured by entropy.

We mathematically describe EntroBank as follows. Assume the capacity of EntropyBank is $N$. We record a sample as $(x,\hat{y}, \alpha, e )$, representing the sample, inferred label, age, and entropy. The inferred label $\hat{y}$ is obtained by $\argmax p(y|x)=\text{softmax}(f_{\theta^T} (x))$. 
The age of a sample starts at 0 and increases over time. The entropy $e$ is calculated as $\sum_y -p(y|x)\log p(y|x)$.

When adding a new sample $(x,\hat{y}, \alpha, e )$ to EntroBank, if memory usage is below capacity, we directly add the sample.  Otherwise, we apply removal strategies in this order: outdated samples, long-persisted over-confident samples, and sample uncertainty.

Considering categorical balance, we make replacements only in dominant classes $\mathcal{D}_c$, defined as classes with the highest sample count based on inferred labels: 
\begin{equation}
    \label{eq:dominant_class}
    \mathcal{D}_c = \{c|m[c]=\max\limits_{i\in[\mathcal{C}]} m[i]\} ,
\end{equation}
where $m[c]$ is the count for each class $c$ indicated by inferred labels $\{\hat{y}\}$, and $[\mathcal{C}]$ is the a set of class indices $1,2,3,...,\mathcal{C}$.  

For timeliness, we define outdated samples as those whose age $\alpha$ exceeds $\mathcal{T}_{forget}$, a hyperparameter:
\begin{equation}
    \label{eq:outdated}
    \mathcal{X}_{od} = \{(x,\hat{y}, \alpha, e )|\alpha \geq \mathcal{T}_{forget}\} .
\end{equation}
We identify outdated samples $\mathcal{X}_{od}$ among the dominated classes $\mathcal{D}_c$. If $\mathcal{X}_{od}$ is not empty, we remove a sample from $\mathcal{X}_{od}$ with the largest age. If $\mathcal{X}_{od}$ is empty, we consider long-persisted over-confident samples.

We define long-persisted over-confident samples as those whose age $\alpha$ exceeds $\mathcal{T}_{mature}$ and have the smallest entropy $e$ among their class in terms of inferred labels. 
 
\begin{equation}
    \label{eq:over_confident}
    \displaystyle
    \mathcal{X}_{oc} = \{(x,\hat{y}, \alpha, e )|\alpha \geq \mathcal{T}_{mature}, e = \min\limits_{\{i|i\in [N],\hat{y}_i=\hat{y}\}} e_i\} ,
\end{equation}
where $[N]$ is a set of samples indices in EntroBank $1,2,3,...,N$. If $\mathcal{X}_{oc}$ is not empty, we remove a sample from $\mathcal{X}_{oc}$ with the lowest entropy. If $\mathcal{X}_{oc}$ is empty, we consider consider sample uncertainty. 

We first define the sample $(x',\hat{y}',\alpha', e')$ with the highest entropy among the dominated classes $\mathcal{D}_c$. If $e<e'$, indicating lower uncertainty, we replace $(x',\hat{y}',\alpha', e')$ with the new sample. Otherwise, we discard the new sample.  The details of the sample adding in EntroBank are outlined in Algorithm \ref{alg:algorithm_memory_bank}.

\begin{algorithm}[!htb]
    \caption{EntroBank for sample adding}
    \label{alg:algorithm_memory_bank}
    \textbf{Input}: sample $x$ and the adapted model $f_{\theta'}$ \\
    \textbf{Define}: dominant classes $\mathcal{D}_c$, outdated samples: $\mathcal{X}_{od}$,\\
    long-existed over-confident samples: $\mathcal{X}_{oc}$.
    
    \begin{algorithmic}[1] 
        \STATE Increase the age for each sample in memory
        \STATE Infer distritbution: $p(y|x)=\text{softmax}(f_{\theta'} (x))$
        \STATE Predicted label: $\hat{y} = \argmax p(y|x)$
        \STATE Initialize age: $\alpha=0$
        \STATE Calculate entropy: $e=\sum_y -p(y|x)\log p(y|x)$
        \IF{memory usage $<$ capacity}
        \STATE Add $(x,\hat{y}, \alpha, e)$ to memory
    
        \ELSE
        \STATE Find $\mathcal{D}_c$ by Eq. \ref{eq:dominant_class}
        \STATE Find $\mathcal{X}_{od}$ by Eq. \ref{eq:outdated} among $\mathcal{D}_c$
        \STATE Find $\mathcal{X}_{oc}$ by Eq. \ref{eq:over_confident} among $\mathcal{D}_c$
        \STATE Find sample $x'$ with highest entropy $e'$ among $\mathcal{D}_c$
        \IF{$\mathcal{X}_{od}$ is not $\varnothing$ }
        \STATE Remove a sample  $\in \mathcal{X}_{od}$ with the largest age
        \STATE Add $(x,\hat{y}, a, e)$ to memory
        \ELSIF{$\mathcal{X}_{oc}$ is not $\varnothing$ }
        \STATE Remove a sample $\in \mathcal{X}_{oc}$ with the lowest entropy
        \STATE Add $(x,\hat{y}, a, e)$ to memory
        \ELSE
        \IF{$e < e' $}
        \STATE Remove the item containing $x'$
        \STATE Add $(x,\hat{y}, a, e)$ to memory
        \ELSE
        \STATE Discard $x$
        \ENDIF
        \ENDIF
        \ENDIF
    \end{algorithmic}
\end{algorithm}

\subsection{Self-Training Adapatation}
\label{sec:self-training}
In the adaptation phase, following the self-training method described in \cite{cotta_cvpr_22}, we employ a teacher-student model to adapt the source model using a self-training loss.  The source pre-trained model is duplicated into a teacher model $f_{\theta'}$ and a student model $f_{\theta}$. At time-step t, we first update the student model parameters from $\theta_t$ to $\theta_{t+1}$ using a self-training loss $\mathcal{L}_s$. This loss applies a strong augmentation view for the student model's input and a weak augmentation view for the teacher model's input \footnote{We adhere to the augmentation practices from prior research \cite{cotta_cvpr_22,rotta_cvpr_23}, utilizing ReSize+CenterCrop for 
the weak augmentation view, and ColorJitter + RandomAffine + RandomHorizontalFlip + GaussianBlur + GaussianNoise as the main augmentations for the strong augmentation view.}: 
\begin{equation}
    \mathcal{L}_{s} = \frac{1}{B}\sum\limits_{i\in[B]} CE(p(\hat{y}|\mathcal{T}_w(x), \theta'_t), p(\hat{y}|\mathcal{T}_s(x), \theta_t)),
\end{equation}

where $CE(\cdot,\cdot)$ represents the cross entropy operation, defined as $CE(p,q) = -\sum p\log q$. Samples $\{x\}$ are drawn from the memory bank, with $B$ denoting the occupation of the memory bank.
$\mathcal{T}_s, \mathcal{T}_w$ denote strong and weak augmentations respectively. $p(\hat{y}|\mathcal{T}_w(x), \theta'_t) = \text{softmax}(f_{\theta'_t}(\mathcal{T}_w(x)))$, where $\theta'_t$ refers to the parameters of the teacher model at time step $t$. $p(\hat{y}|\mathcal{T}_s(x), \theta_t) = \text{softmax}(f_{\theta_t}(\mathcal{T}_s(x)))$, where $\theta_t$ refers to the parameters of the student model at time step $t$.

The teacher model is updated using an exponential moving average of the parameters of the student model: 
\begin{equation}
    \theta'_{t+1} = (1-\nu_m) \theta'_{t} + \nu\theta_{t+1} ,
\end{equation}
where $\nu_m$ is a parameter controlling the update rate in the teacher model. 

In summary, we utilize EntroBank to manage sample storage within capacity constraints and replace batch normalization with ResiBN. We periodically adapt the source pre-trained model through a self-training approach on the memory samples, incorporating a soft alignment loss on batch normalization. 

\section{Experiments}
In this section, we design experiments to compare \method with state-of-the-art (SOTA) methods under practical test-time adaptation (PTTA) setting \cite{rotta_cvpr_23} on CIFAR10-C, CIFAR100-C, and ImageNet-C \cite{robustness_benchmark_ICLR_2019}. We then conduct ablation studies of \method on CIFAR100-C to demonstrate the effectiveness of each module design. Finally, we perform parameter analysis on the parameters $\eta_t$, $\mathcal{T}_{forget}$ and $\mathcal{T}_{mature}$.

\subsection{Datasets}
We validate our method on CIFAR10C, CIFAR100-C, and ImageNet-C, which are benchmarks created by \cite{robustness_benchmark_ICLR_2019}. CIFAR10-C and CIFAR100-C are corruption versions of CIFAR10 and CIFAR100 \cite{cifar10_100_2009}, featuring 15 types of corruption, each with 5 different degrees of severity. Each type and degree of corruption in CIFAR10-C and CIFAR100-C contains 10,000 samples, falling into 10 and 100 classes, respectively. ImageNet-C is a corruption version of the ImageNet \cite{imagenet} validation dataset, with 15 types of corruption and 5 different degrees of severity. Each type and degree of corruption in ImageNet-C contains 5,000 samples falling into 1,000 classes.

\subsection{Baselines}
We compare our method with the following test-time adaptation algorithms:
1) \textbf{Source}: This approach infers test-time samples using the pre-trained model without any adaptation updates; 2) Prediction time batch normalization (\textbf{BN}) \cite{test_time_batch_norm}: This method freezes the pre-trained model weights and infers test-time samples by using the batch norm statistics from the test batch; 3) Pseudo-labeling (\textbf{PL}) \cite{pseudo_labeling}: This technique generates pseudo labels using the inferred labels and updates models using the cross entropy on the pseudo labels; 4) \textbf{TENT} \cite{tent_iclr_21}: This method aims to adapt model by minimizing the entropy of predictions on test data to reduce generalization error; 5) \textbf{LAME} \cite{lame_cvpr_22}: This approach fixes the pre-trained model but adjusts the output probability by adding Laplacian constraints to the local batch; 6) \textbf{CoTTA} \cite{cotta_cvpr_22}: This algorithm uses a teacher-student model with an average augmentation strategy and stochastic weight recovery; 7) \textbf{NOTE} \cite{note_neurips_22}: This method employs an instance-aware batch normalization and a category-balanced memory bank.  8) \textbf{RoTTA} \cite{rotta_cvpr_23}: This method maintains global batch norm statistics with exponential average updates on test-batch statistics and a class-balanced memory bank considering uncertainty and timeliness. RoTTA adapts the pre-trained model by a timeliness-aware adaptation loss with a teacher-student model. 

\begin{table*}[h]
    \centering
    \caption{Classification error rate (\%) of the task CIFAR10 $\to$ CIFAR10-C online continual test-time adaptation evaluated on WideResNet-28 at the largest corruption severity 5. Samples in each corruption are correlatively sampled under the setup \setting. 
    }
    \label{table:cifar10}
    \VspaceBefore
    \resizebox{\textwidth}{!}{
    \renewcommand{\arraystretch}{0.8}
    {
    \begin{tabular}{l|ccccccccccccccc|c}
        \toprule[1.2pt]
        Time & \multicolumn{15}{l|}{$t\xrightarrow{\hspace*{18.5cm}}$}& \\ 
        \hline
        Method & \rotatebox[origin=c]{45}{motion} & \rotatebox[origin=c]{45}{snow} & \rotatebox[origin=c]{45}{fog} & \rotatebox[origin=c]{45}{shot} & \rotatebox[origin=c]{45}{defocus} & \rotatebox[origin=c]{45}{contrast} & \rotatebox[origin=c]{45}{zoom} & \rotatebox[origin=c]{45}{brightness} & \rotatebox[origin=c]{45}{frost} & \rotatebox[origin=c]{45}{elastic} & \rotatebox[origin=c]{45}{glass} & \rotatebox[origin=c]{45}{gaussian} & \rotatebox[origin=c]{45}{pixelate} & \rotatebox[origin=c]{45}{jpeg} & \rotatebox[origin=c]{45}{impulse}
        & Avg. \\ 
        
        \midrule
        Source & 34.8 & 25.1 & 26.0 & 65.7 & 46.9 & 46.7 & 42.0 & 9.3 & 41.3 & 26.6 & 54.3 & 72.3 & 58.5 & 30.3 & 72.9 & 43.5 \\ 
        BN~\cite{test_time_batch_norm} & 73.2 & 73.4 & 72.7 & 77.2 & 73.7 & 72.5 & 72.9 & 71.0 & 74.1 & 77.7 & 80.0 & 76.9 & 75.5 & 78.3 & 79.0 & 75.2 \\ 
        PL~\cite{pseudo_labeling} & 73.9 & 75.0 & 75.6 & 81.0 & 79.9 & 80.6 & 82.0 & 83.2 & 85.3 & 87.3 & 88.3 & 87.5 & 87.5 & 87.5 & 88.2 & 82.9 \\ 
        TENT~\cite{tent_iclr_21} & 74.3 & 77.4 & 80.1 & 86.2 & 86.7 & 87.3 & 87.9 & 87.4 & 88.2 & 89.0 & 89.2 & 89.0 & 88.3 & 89.7 & 89.2 & 86.0 \\ 
        LAME~\cite{lame_cvpr_22} & 29.5 & \bf 19.0 & 20.3 & 65.3 & 42.4 & 43.4 & 36.8 & 5.4 & 37.2 & \bf 18.6 & 51.2 & 73.2 & 57.0 & \bf 22.6 & 71.3 & 39.5 \\ 
        CoTTA~\cite{cotta_cvpr_22} & 77.1 & 80.6 & 83.1 & 84.4 & 83.9 & 84.2 & 83.1 & 82.6 & 84.4 & 84.2 & 84.5 & 84.6 & 82.7 & 83.8 & 84.9 & 83.2 \\ 
        NOTE~\cite{note_neurips_22} & \bf 18.0 & 22.1 & 20.6 & 35.6 & 26.9 & \underline{13.6} & 26.5 & \underline{17.3} & 27.2 & 37.0 & 48.3 & 38.8 & 42.6 & 41.9 & 49.7 & 31.1 \\ 
        RoTTA~\cite{rotta_cvpr_23} & \underline{18.1} & 21.3 & \underline{18.8} & \underline{33.6} & \bf 23.6 & 16.5 & \underline{15.1} & \bf 11.2 & \underline{21.9} & 30.7 & \underline{39.6} & \underline{26.8} & \underline{33.7} & 27.8 & \underline{39.5} & \underline{25.2} \\ 
        \midrule
        \method & 18.4 & \underline{19.5} & \bf 15.5 & \bf 30.5 & \underline{23.8} & \bf 12.2 & \bf 14.0 & 9.3 & \bf 18.5 & \underline{24.6} & \bf 35.8 & \bf 24.9 & \bf 27.7 & \bf 22.6 & \bf 39.1 & \bf 22.4 \\ 
    \bottomrule[1.2pt]
    \end{tabular}
    }
    }
    \VspaceAfter
\end{table*}

\begin{table*}[!ht]
    \centering
    \caption{    Classification error rate (\%) of the task CIFAR100 $\to$ CIFAR100-C online continual test-time adaptation evaluated on the ResNeXt-29 architecture at the largest corruption severity 5. Samples in each corruption are correlatively sampled under the setup \setting. 
    }
    \label{table:cifar100}
    \VspaceBefore
    \resizebox{\linewidth}{!}{
    \renewcommand{\arraystretch}{0.8}
    {
    \begin{tabular}{l|ccccccccccccccc|c}
        \toprule[1.2pt]
        Time & \multicolumn{15}{l|}{$t\xrightarrow{\hspace*{18.5cm}}$}& \\ \hline
        Method & \rotatebox[origin=c]{45}{motion} & \rotatebox[origin=c]{45}{snow} & \rotatebox[origin=c]{45}{fog} & \rotatebox[origin=c]{45}{shot} & \rotatebox[origin=c]{45}{defocus} & \rotatebox[origin=c]{45}{contrast} & \rotatebox[origin=c]{45}{zoom} & \rotatebox[origin=c]{45}{brightness} & \rotatebox[origin=c]{45}{frost} & \rotatebox[origin=c]{45}{elastic} & \rotatebox[origin=c]{45}{glass} & \rotatebox[origin=c]{45}{gaussian} & \rotatebox[origin=c]{45}{pixelate} & \rotatebox[origin=c]{45}{jpeg} & \rotatebox[origin=c]{45}{impulse}
        & Avg. \\ 
        
        \midrule
        Source & 30.8 & 39.5 & 50.3 & 68.0 & 29.3 & 55.1 & 28.8 & 29.5 & 45.8 & 37.2 & 54.1 & 73.0 & 74.7 & 41.2 & \underline{39.4} & 46.4 \\ 
        BN~\cite{test_time_batch_norm} & 48.5 & 54.0 & 58.9 & 56.2 & 46.4 & 48.0 & 47.0 & 45.4 & 52.9 & 53.4 & 57.1 & 58.2 & 51.7 & 57.1 & 58.8 & 52.9 \\ 
        PL~\cite{pseudo_labeling} & 50.6 & 62.1 & 73.9 & 87.8 & 90.8 & 96.0 & 94.8 & 96.4 & 97.4 & 97.2 & 97.4 & 97.4 & 97.3 & 97.4 & 97.4 & 88.9 \\ 
        TENT~\cite{tent_iclr_21} & 53.3 & 77.6 & 93.0 & 96.5 & 96.7 & 97.5 & 97.1 & 97.5 & 97.3 & 97.2 & 97.1 & 97.7 & 97.6 & 98.0 & 98.3 & 92.8 \\ 
        LAME~\cite{lame_cvpr_22} & \bf 22.4 & \bf 30.4 & 43.9 & 66.3 & \bf 21.3 & 51.7 & \bf 20.6 & \bf 21.8 & 39.6 & \bf 28.0 & 48.7 & 72.8 & 74.6 & \bf 33.1 & \bf 32.3 & 40.5 \\ 
        CoTTA~\cite{cotta_cvpr_22} & 49.2 & 52.7 & 56.8 & 53.0 & 48.7 & 51.7 & 49.4 & 48.7 & 52.5 & 52.2 & 54.3 & 54.9 & 49.6 & 53.4 & 56.2 & 52.2 \\ 
        NOTE~\cite{note_neurips_22} & 45.7 & 53.0 & 58.2 & 65.6 & 54.2 & 52.0 & 59.8 & 63.5 & 74.8 & 91.8 & 98.1 & 98.3 & 96.8 & 97.0 & 98.2 & 73.8 \\ 
        RoTTA~\cite{rotta_cvpr_23}  & 31.8 & 36.7 & \underline{40.9} & \underline{42.1} & 30.0 & \underline{33.6} & 27.9 & 25.4 & \underline{32.3} & 34.0 & \underline{38.8} & \underline{38.7} & \bf 31.3 & 38.0 & 42.9 & \underline{35.0} \\ 
        \midrule
        \method & \underline{29.2} & \underline{33.9} & \bf 39.5 & \bf 39.4 & \underline{28.4} & \bf 29.2 & \underline{26.5} & \underline{24.8} & \bf 30.8 & \underline{33.9} & \bf 37.5 & \bf 38.6 & \underline{31.6} & \underline{37.9} & 41.5 & \bf 33.5 \\ 
    \bottomrule[1.2pt]
    \end{tabular}
    }
    }
    \VspaceAfter
\end{table*}

\subsection{Implementation Details}
We follow the model selection criteria of previous methods \cite{cotta_cvpr_22,rotta_cvpr_23} from the RobustBench benchmark \cite{robust_benchmark_NIPS_21}, using WildResNet-28 \cite{WideResidualNetworks_BMVC_2016} for CIFAR10 $\rightarrow$  CIFAR10-C, ResNeXt-29 \cite{ResNeXt_CVPR_2017} for CIFAR100 $\rightarrow$ CIFAR100-C and the standard resnet50 from \cite{robust_benchmark_NIPS_21} for ImageNet $\rightarrow$ ImageNet-C. 
Consistent with the dataset protocol in \cite{rotta_cvpr_23}, we simulate continuous distribution changes by altering the test domain at severity level 5 and use Dirichlet distribution sampling to model a non i.i.d. test stream for CIFAR10-C/100-C.
 Due to the limited number of samples per class in ImageNet-C, Dirichlet sampling cannot form an effective temporal correlation. Following \cite{note_neurips_22}, we sort ImageNet-C classes within each corruption type to mimic temporal correlation, with further details provided in the appendix. We use Adam as the optimizer with a learning rate of 1.0 $\times$ 10$^\text{-3}$ and a beta value of 0.9. The batch size is set to 64. The memory bank capacity and the update frequency are also set to 64 for all memory bank approaches for a fair comparison. The parameters $\nu_m=1.0\times 10^{-3}$, $\nu_b=0.05$, and $\delta=0.1$ are chosen. We use $\eta_t$ values of $0.01$ for CIFAR10-C/100-C and $0.05$ for ImageNet-C, with the larger $\eta_t$ for ImageNet-C accounting for dataset complexity, which validated in our parameter analysis in sec. \ref{sec:para_analysis}.

\subsection{Main Results}

\paragraph{CIFAR10 $\rightarrow$ CIFAR10-C}

We report the classification error rate for the CIFAR-10-to-CIFAR10C task in Table \ref{table:cifar10}. \method achieves the lowest average error value and outperforms other methods in 10 out of 15 corruption types. When compared to the best-performing baseline, RoTTA \cite{rotta_cvpr_23}, our method reduces the average error rate from 25.2\% to 22.4\%.  
NOTE \cite{note_neurips_22} and RoTTA \cite{rotta_cvpr_23} obtain results comparable to ours for the initial corruption type, but they exhibit weaker performance on subsequent corruption types. This trend highlights the effectiveness of ResiBN in reducing parameter overfitting. Additionally, our method consistently matches or exceeds the Source model across all corruption types, indicating its ability to mitigate potential model degradation

\paragraph{CIFAR100 $\rightarrow$ CIFAR100-C} 

We report the performance of \method on the CIFAR100-to-CIFAR100-C task in Table \ref{table:cifar100}. \method achieves the lowest average classification error, reducing the average error rate from 35.0\% to 33.5\% in comparison to the previously best-performing method, RoTTA \cite{rotta_cvpr_23}. LAME \cite{lame_cvpr_22}, a non-parametric approach, yields results comparable to \method and achieves the best performance in several corruption types. However, the performance of LAME varies significantly across corruption types, largely depending on the performance of the pre-trained model.

\paragraph{ImageNet $\rightarrow$ ImageNet-C} 

We report the results of all baselines and \method on the ImageNet-to-ImageNet-C task in Table \ref{table:imagenet}. \method outperforms other methods in average classification error and achieves the best performance in 13 out of 15 corruption types. It reduces the average test error from 71.6\% to 66.7\% compared to RoTTA \cite{rotta_cvpr_23}. Specifically, in the corruption types \textit{jpeg} and \textit{impulse} among the last two domains, \method decreases the classification error from 58.8\% to 53.6\% and from 84.5\% to 75.5\% respectively, compared with the previous best results. These findings demonstrate the effectiveness \method in reducing overfitting and model degradation

\begin{table*}[!ht]
    \centering
    \caption{Classification error rate (\%) of the task ImageNet $\to$ ImageNet-C online continual test-time adaptation evaluated on the resnet50 architecture at the largest corruption severity 5. Samples in each corruption are correlatively arranged following \protect\cite{note_neurips_22}
    }
    \label{table:imagenet}
    \VspaceBefore
    \resizebox{\linewidth}{!}{
    \renewcommand{\arraystretch}{0.8}
    {
    \begin{tabular}{l|ccccccccccccccc|c}
        \toprule[1.2pt]
        Time & \multicolumn{15}{l|}{$t\xrightarrow{\hspace*{18.5cm}}$}& \\ \hline
        Method & \rotatebox[origin=c]{45}{motion} & \rotatebox[origin=c]{45}{snow} & \rotatebox[origin=c]{45}{fog} & \rotatebox[origin=c]{45}{shot} & \rotatebox[origin=c]{45}{defocus} & \rotatebox[origin=c]{45}{contrast} & \rotatebox[origin=c]{45}{zoom} & \rotatebox[origin=c]{45}{brightness} & \rotatebox[origin=c]{45}{frost} & \rotatebox[origin=c]{45}{elastic} & \rotatebox[origin=c]{45}{glass} & \rotatebox[origin=c]{45}{gaussian} & \rotatebox[origin=c]{45}{pixelate} & \rotatebox[origin=c]{45}{jpeg} & \rotatebox[origin=c]{45}{impulse}
        & Avg. \\ 
        
        \midrule
        Source & 85.5 & 83.5 & 76.4 & 96.5 & 81.8 & 94.6 & 78.1 & 41.2 & 77.4 & 83.0 & 89.7 & 97.1 & 79.1 & 67.5 & 97.5 & 81.9 \\ 
        BN~\cite{test_time_batch_norm} & 86.0 & 80.8 & 71.1 & 90.4 & 92.4 & 90.4 & 79.0 & 58.4 & 81.9 & 75.5 & 92.4 & 90.7 & 71.1 & 76.2 & 90.6 & 81.8 \\ 
        PL~\cite{pseudo_labeling} & 87.2 & 88.3 & 86.4 & 97.2 & 98.9 & 99.6 & 99.1 & 98.5 & 99.6 & 99.6 & 99.7 & 99.8 & 99.7 & 99.7 & 99.7 & 96.9 \\ 
        TENT~\cite{tent_iclr_21} & 84.9 & 78.9 & 69.4 & 87.9 & 89.4 & 87.6 & 79.1 & 64.4 & 83.0 & 77.2 & 91.2 & 88.9 & 78.0 & 80.4 & 89.6 & 82.0 \\ 
        LAME~\cite{lame_cvpr_22} & 85.3 & 83.4 & 76.1 & 96.7 & \bf 80.6 & 94.1 & 76.1 & \bf 34.6 & 76.2 & 83.2 & 89.9 & 97.4 & 77.5 & 63.8 & 97.9 & 80.8 \\ 
        CoTTA~\cite{cotta_cvpr_22} & 86.9 & 80.4 & 71.7 & \underline{87.3} & 90.8 & 89.4 & 78.1 & 60.5 & 78.8 & 73.1 & 89.4 & \underline{85.0} & 67.8 & 72.6 & \underline{84.5} & 79.8 \\ 
        NOTE~\cite{note_neurips_22} & 83.9 & 79.6 & 78.2 & 93.5 & 97.3 & 94.4 & 96.7 & 92.1 & 97.1 & 98.1 & 99.3 & 99.4 & 98.6 & 98.9 & 99.6 & 93.8 \\ 
        RoTTA~\cite{rotta_cvpr_23} & \underline{78.3} & \underline{76.2} & \underline{59.9} & 88.2 & 87.7 & \underline{84.7} & \underline{66.1} & 40.4 & \underline{69.6} & \underline{58.4} & \underline{78.1} & 86.2 & \underline{56.7} & \underline{58.8} & 85.6 & \underline{71.6} \\ 
        \midrule
        \method & \bf 74.4 & \bf 69.2 & \bf 54.0 & \bf 82.9 & \underline{80.7} & \bf 75.8 & \bf 63.9 & \underline{37.5} & \bf 65.6 & \bf 57.2 & \bf 76.8 & \bf 77.5 & \bf 55.7 & \bf 53.6 & \bf 75.5 & \bf 66.7 \\ 

    \bottomrule[1.2pt]
    \end{tabular}
    }
    }
    \VspaceAfter
\end{table*}

\subsection{Ablation Study}

In Table \ref{table:ablation}, we present different model variants by replacing or removing components to validate the module effectiveness in \method in the CIFAR100-to-CIFAR100-C task.  In row \rownumber{2}, We first replace ResiBN by prediction time batch normalization (BN) \cite{test_time_batch_norm}, resulting in a significant increase in average test error from 33.5 \% to 51.6 \%, due to unstable batch-wise statistics. Substituting ResiBN with RBN \cite{rotta_cvpr_23} led to an increase in average error from 33.5 \% to 35.4 \% (row \rownumber{3}). RBN uses an exponential moving average update for target statistics during inference but does not address the potential for overfitting on the test batches. Next, we individually remove three strategies in EntroBank. Eliminating the strategy for outdated samples (row \rownumber{4}) raises the error rate to 33.9 \%. Omitting the strategy for long-persisted over-confident samples increases the error rate to 35.4 \% (row \rownumber{5}). Removing the strategy for sample uncertainty significantly elevated the error rate to 38.2 \% (row \rownumber{6}). These results demonstrate the importance of the three strategies in EntroBank.

\begin{table}[!ht]
  \centering
	\caption{
    \textbf{Important component for \method.} Average classification error rate (\%) of task CIFAR100 $\to$ CIFAR100-C. 
    For the different variants, we \colorbox{Light}{highlight} the differences from the default \method setting. 
  }
  \label{table:ablation}
\small
  \setlength{\tabcolsep}{4pt}
  \VspaceBefore
  \renewcommand{\arraystretch}{0.8}
  \begin{tabular}{@{}p{.8em}@{}l ccccccc@{}}
    \toprule[1pt]
        &Method & AdBN & RBN & ResiBN & $\mathcal{X}_{od}$ & $\mathcal{X}_{oc}$ & SU & CIFAR100C \\
            \midrule 
        \rownumber{1} & \method & \xmark & \xmark & \checkmark & \checkmark & \checkmark & \checkmark & 33.5 \\ 
        \rownumber{2} &  & \colorbox{Light}{\checkmark} & \xmark & \colorbox{Light}{\xmark} & \checkmark & \checkmark & \checkmark & 51.6 \\ 
        \rownumber{3} &  & \xmark & \colorbox{Light}{\checkmark} & \colorbox{Light}{\xmark} & \checkmark & \checkmark & \checkmark & 35.4 \\ 
        \rownumber{4} &  & \xmark & \xmark & \checkmark & \colorbox{Light}{\xmark} & \checkmark & \checkmark & 33.9 \\ 
        \rownumber{5} &  & \xmark & \xmark & \checkmark & \checkmark & \colorbox{Light}{\xmark} & \checkmark & 35.4 \\ 
        \rownumber{6} &  & \xmark & \xmark & \checkmark & \checkmark & \checkmark & \colorbox{Light}{\xmark} & 38.2 \\ 
    \bottomrule[1pt]
\multicolumn{9}{l}{$\mathcal{X}_{od}$: outdated samples, $\mathcal{X}_{oc}$: long-existed over-confident samples}\\
\multicolumn{9}{l}{SU: sample uncertainty, ResiBN: Resilient Batch Norm }\\
\multicolumn{9}{l}{ BN: prediction time batch normalization \cite{test_time_batch_norm} }\\
\multicolumn{9}{l}{ RBN: robust Batch Norm \cite{rotta_cvpr_23}}\\

  \end{tabular}
  \VspaceAfter
\label{tab:components}
\end{table}

\subsection{Parameter Analysis}
\label{sec:para_analysis}
In this section, we examine the sensitivity of hyperparameters: $\eta_t$, $\mathcal{T}_{mature}$, $ \mathcal{T}_{forget}$ by adjusting each parameter from the default setup. For each parameter, we calculate the average classification error across the CIFAR10-to-CIFAR10-C, the CIFAR100-to-CIFAR100-C, and ImageNet-to-ImageNet-C tasks over 15 corruptions types with severity 5 using the default corruption order and correlation sampling methods. 

First, we investigate the parameter $\eta_t$ in Table \ref{table:paras_eta_t}. This parameter determines the strength of soft alignments between source and target statistics in batch normalization. We vary $\eta_t$ within [0.001, 0.005, 0.01, 0.05, 0.1]. For CIFAR10-C and CIFAR100-C, the optimal results occur at $\eta_t =0.01$. Deviations from this value lead to performance decline. In contrast, the best result of ImageNet-C is achieved at $\eta_t = 0.05$, differing from  CIFAR10-C/100-C due to the complexity of ImageNet and the large number of classes. This data complexity makes capturing batch normalization statistics in test time challenging and prone to overfitting. Consequently, \method requires stronger soft alignments of batch normalization statistics for the ImageNet-to-ImageNet-C task. 
\begin{table}[!ht]
    \centering
    \caption{\textbf{Parameter analysis on $\eta_t$.} Average classification error (\%).
    }
    \label{table:paras_eta_t}
    \setlength{\tabcolsep}{3pt}
    \small
    \renewcommand{\arraystretch}{0.8}
    \VspaceBefore
    \begin{tabular}{lccc}
    \toprule[1pt]
        $\eta_t$ & CIFAR10-C & CIFAR100-C & ImageNet-C \\ 
         \midrule 
        0.001 & 23.0 & 34.1 & 73.5 \\ 
        0.005 & 22.7 & 33.8 & 72.0 \\ 
        0.01 & 22.4 & 33.5 & 70.6 \\ 
        0.05 & 25.6 & 33.9 & 66.7 \\ 
        0.1 & 30.6 & 35.8 & 68.4 \\ 
        \bottomrule[1pt]
    \end{tabular}
    \VspaceAfter
\end{table}

Next, we examine $\mathcal{T}_{mature}$ in Table \ref{table:paras_mature_age}. This parameter regulates the duration low-entropy samples remain.  Increasing $\mathcal{T}_{mature}$ from 200 leads to a gradual increase in error rate across all datasets: from 22.4 \% to 23.4 \% on CIFAR10-C, from 33.5 \% to 34.9 \% on CIFAR100-C, and 66.7 \% to 67.8 \% on ImageNet-C. This indicates that extending the existence of over-confident samples declines performance.

\begin{table}[!ht]
    \centering
        \caption{\textbf{Parameter analysis on $\mathcal{T}_{mature}$.} Average classification error (\%).
    }
    \label{table:paras_mature_age}
    \small
    \VspaceBefore
    \renewcommand{\arraystretch}{0.8}
    \begin{tabular}{lccc}
    \toprule[1pt]
    $\mathcal{T}_{mature}$ & CIFAR10-C & CIFAR100-C & ImageNet-C \\ 
     \midrule 
        100 & 22.8 & 33.2 & 67.1 \\ 
        200 & 22.4 & 33.5 & 66.7 \\ 
        300 & 23.0 & 33.6 & 67.1 \\ 
        400 & 23.1 & 34.0 & 67.1 \\ 
        500 & 23.1 & 34.2 & 67.6 \\ 
        600 & 23.0 & 34.4 & 67.6 \\ 
        700 & 23.1 & 34.8 & 67.8 \\ 
        800 & 23.4 & 34.9 & 67.8 \\ 
    \bottomrule[1pt]
    \end{tabular}
    \VspaceAfter
\end{table}

Lastly, we analyze $\mathcal{T}_{forget}$ in Table $\ref{table:paras_forget_age}$. We find that \method is relatively insensitive to $\mathcal{T}_{forget}$. This is because a small proportion of samples become outdated. Most samples in the memory bank are updated in prediction probability $p(\hat{y}|x,\theta')$, resulting in a gradual decrease in entropy. Additionally, $\mathcal{T}_{forget}$ being significantly larger than $ \mathcal{T}_{mature}$ means most samples are classified as long-persisted over-confident samples before being considered outdated.

\begin{table}[!ht]
    \centering
            \caption{\textbf{Parameter analysis on $\mathcal{T}_{forget}$.}   Average classification error (\%).
    }
    \label{table:paras_forget_age}
    \small
    \VspaceBefore
    \renewcommand{\arraystretch}{0.8}
    \begin{tabular}{lccc}
    \toprule[1pt]
        $\mathcal{T}_{forget}$ & CIFAR10-C & CIFAR100-C & ImageNet-C \\ 
         \midrule 
        500 & 22.6 & 33.6 & 66.9 \\ 
        1000 & 22.4 & 33.5 & 66.7 \\ 
        1500 & 22.2 & 33.7 & 67.1 \\ 
        2000 & 22.5 & 33.6 & 66.8 \\ 
        2500 & 22.4 & 33.8 & 66.9 \\ 
    \bottomrule[1pt]
    \end{tabular}
    \VspaceAfter
\end{table}

\section{Conclusion}
Addressing potential model degradation in a practical test-time adaptation setting, we propose a resilient practical test-time adaptation (\method) method, which incorporates three parts: resilient batch normalization, entropy-driven memory bank, and self-training adaptation. Resilient batch normalization updates global target statistics progressively with test-batch statistics in batch normalization and implements soft alignment on these target statistics by minimizing to reduce parameter overfitting.  The entropy-driven memory bank is developed to provide high-quality samples, considering three aspects: timeliness, the persistence of over-confident samples, and sample uncertainty. The model is adapted periodically by a teacher-student model using a self-training loss on memory samples, incorporating soft alignment losses. Extensive experiments, ablation studies, and parameter analysis are conducted for validation of the effectiveness of \method. 

\bibliographystyle{named}
\bibliography{ijcai24}


\appendix
\clearpage

\begin{center}
    \textbf{\Large APPENDIX}
\end{center}
\vspace{3mm}

In this appendix, we further explain the technical details of \method including divergence selection in sec. \ref{sec:divergence_selection}, implementation details in sec. \ref{sec:implementation details}, and extra experiment results in sec. \ref{sec:extra experiment results}.

\section{Divergence Selection}
\label{sec:divergence_selection}
In sec. \ref{sec:resilient_BN}, we choose Wasserstein distance for soft alignments on batch normalization statistics $\mu_t, \sigma_t$, and we illustrate the reason why not using KL or JS divergences to maintain numerical stability. 

We first obtain the KL divergence between source and target statistics:
\begin{align}
    KL&(\mathcal{N}(\mu_t, \sigma_t^2), \mathcal{N}(\mu_s,\sigma_s^2))=  \nonumber\\
    & \frac{1}{2}(\frac{\sigma_t^2}{\sigma_s^2} + \frac{(\mu_t - \mu_s)^2}{\sigma_s^2} - 1+2\ln(\frac{\sigma_s}{\sigma_t})).
\end{align}
Taking derivative towards $\mu_t,\sigma_t^2, \sigma_t$, we have:
\begin{align}
    \frac{dKL(\cdot)}{d\mu_t} &= \frac{(\mu_t - \mu_s)}{\sigma_s^2},\\
    \frac{dKL(\cdot)}{d\sigma_t^2} &=   \frac{1}{2\sigma_s^2}-\frac{1}{2\sigma_t^2},  \\
    \frac{dKL(\cdot)}{d\sigma_t} &=\frac{\sigma_t^2 -\sigma_s^2}{\sigma_t \sigma_s^2}.\\
\end{align}
We observe the denominators of all derivatives contain either  $\sigma_s$ or $\sigma_t$, contributing to numerical stability. We can obtain the same conclusion for the case  $KL(\mathcal{N}(\mu_s,\sigma_s^2),\mathcal{N}(\mu_t, \sigma_t^2))$ using the same approach. 

We now turn to JS divergence between source and target statistics:
\begin{align}
       JS&(\mathcal{N}(\mu_t, \sigma_t^2), \mathcal{N}(\mu_s,\sigma_s^2))=  \nonumber\\
    & \frac{1}{4} \left( \frac{\sigma_t^2}{\sigma_s^2} + \frac{\sigma_s^2}{\sigma_t^2} +(\frac{1}{\sigma_s^2}+\frac{1}{\sigma_t^2})(\mu_t - \mu_s)^2 -2 \right).
\end{align}
We take derivatives towards $\mu_t,\sigma_t^2, \sigma_t$:
\begin{align}
    \frac{dJS(\cdot)}{d\mu_t} &=(\frac{1}{2\sigma_s^2}+\frac{1}{2\sigma_t^2})(\mu_t - \mu_s) ,\\
    \frac{dJS(\cdot)}{d\sigma_t^2} &=   \frac{1}{4\sigma_s^2} - \frac{\sigma_s^2}{4\sigma_t^4}+\frac{(\mu_t-\mu_s)^2}{4 \sigma_t^4},  \\
    \frac{dJS(\cdot)}{d\sigma_t} &=\frac{\sigma_t}{2\sigma_s^2} - \frac{\sigma_s^2}{2\sigma_t^3}+\frac{(\mu_t-\mu_s)^2}{2 \sigma_t^3}.\\
\end{align}
Still, either $\sigma_t$ or $\sigma_s$ occurs in the denominators as a factor, leading to numerical instability. This proved why we can not use JS or KL divergence for soft alignments in batch normalization. 

\section{Implementation Details}
\label{sec:implementation details}
\subsection{Detail Parameter Setting}
We set the following default parameters: 
\begin{itemize}
    \item $\mathcal{T}_{forget} = 1,000.$
    \item $\mathcal{T}_{mature} = 200.$
    \item Augmentation orders in CIFAR10-C, CIFAR100-C and ImageNet-C follow \cite{rotta_cvpr_23}.
    \item In Dirichlet sampling for CIFAR10-C and CIFAR100-C, we choose the number of time slots as same as the number of classes and $\delta=0.1$. Detail implementation is shown in sec.  \ref{sec:correlation_sampling}
    \item The severity is 5 for all corruption types. 
    \item A fix random seed = $1$ for all experiments .
\end{itemize}
We run all experiments on one V100 GPU. 

\subsection{Correlation Sampling}
\label{sec:correlation_sampling}
\textbf{Dirichlet Sampling in CIFAR10-C/100-C}
Traditional test-time adaptation assumes independent identical distributed (i.i.d) test samples, while the test samples might be highly correlated and thus be non-i.i.d. We follow \cite{rotta_cvpr_23} in correlation sampling setting in CIFAR10-C and CIFAR100-C tasks. For each corruption type, we set $K$ time slots. We assume each time slot given a class satisfies a categorical distribution $Cat(\pi)$ which approximately satisfies a Dirichlet distribution$Dir(\delta,\delta,...,\delta)$,
\begin{align}
\pi \sim \text{Dir} (\delta,\delta,...,\delta),\\
    p(T=k|y=c) \sim \text{Cat}(\pi),   
\end{align}

where $p(T=k|y=c)$ represents the probability of sample belonging to time slot $k$ given class $c$, and $T$ denotes the time slot variable. 
After ensuring $p(T|y)$ for each time slot and each class, we randomly samples of each class to each time slot according to $p(T|y)$. Finally, we shuffle the samples at each time slot. In this way, we mimic the temporal correlation at different time slots. In the experiment, we set the number of time slots $K$ equal to the number of classes $C$ and $\delta = 0.1$. 

Notice that, if $\delta$ is large, then $p(T=k|y=c) \sim \text{Cat}(\pi)\approx \text{Cat}(\frac{1}{C},\frac{1}{C},...,\frac{1}{C})$. In such a scenario, each class is evenly distributed in each time slot, degenerating to i.i.d. test samples. If $\delta$ is small, then $p(T=k|y=c) \sim \text{Cat}(\pi)$ is highly imbalanced, and the temporal correlation can thus be simulated. 

\textbf{Correlated Sampling in ImageNet-C} Each class contains only five samples in each corruption type, making Dirichlet sampling unavailable for correlation simulation. We thus follow \cite{note_neurips_22} to sort the ImageNet-C by the labels to simulate correlation. Specifically, the test samples are arranged by their labels and samples in each class will consecutively appear at inference. 

\section{Extra Experiment Results}
\label{sec:extra experiment results}
This section reports parameter analysis on the $\nu_b$, $\nu_m$, and learning rate. For each parameter, we calculate the average classification error across the CIFAR10-to-CIFAR10-C, the CIFAR100-to-CIFAR100-C, and ImageNet-to-ImageNet-C tasks over 15 corruptions types with severity 5 using the default corruption order and correlation sampling methods.

We first explore the influence from $\nu_b$ by running experiments with $\nu_b$ in [0.001, 0.005, 0.01, 0.05, 0.1, 0.5]. $\nu_b$ represents the extent to which the global target statistics are updated. Low $\nu_b$ exhibits large performance declines as the target statistics are almost unchanged and the model suffers from domain shifts. 
High $\nu_b$ shows relatively smaller performance declines.

\begin{table}[!ht]
    \centering
        \caption{\textbf{Parameter analysis on $\nu_b$.} Average classification error (\%).
    }
    \label{table:paras_batch_norm_ema_nu}
    \small
    \VspaceBefore
    \renewcommand{\arraystretch}{0.8}
    \begin{tabular}{lccc}
    \toprule[1pt]
    $\nu_b$ & CIFAR10-C & CIFAR100-C & ImageNet-C \\ 
     \midrule 
        0.001 & 36.6  & 56.2  & 80.7  \\ 
        0.005 & 28.6  & 40.6  & 76.2  \\ 
        0.01 & 25.2  & 35.4  & 73.1  \\ 
        0.05 & 22.4  & 33.5  & 66.7  \\ 
        0.1 & 23.7  & 34.3  & 66.9  \\ 
        0.5 & 28.3  & 35.6  & 68.8 \\ 
    \bottomrule[1pt]
    \end{tabular}
    \VspaceAfter
\end{table}

Next, we investigate the impact of $\nu_m$ by using a range [0.0001, 0.0005, 0.001, 0.005, 0.01, 0.05].  $\nu_m$ represents how fast the teacher model is updated. Large $\nu_m$ results in a significant performance degradation. For instance, the average class error increases from 22.4 \% to 38.1 \% in the CIFAR10-C task when raising $\nu_m$ from 0.001 to 0.05. However, the model is non-sensitive to lower $\nu_m$.

\begin{table}[!ht]
    \centering
        \caption{\textbf{Parameter analysis on $\nu_m$.} Average classification error (\%).
    }
    \label{table:paras_model_ema_nu}
    \small
    \VspaceBefore
    \renewcommand{\arraystretch}{0.8}
    \begin{tabular}{lccc}
    \toprule[1pt]
    $\nu_m$ & CIFAR10-C & CIFAR100-C & ImageNet-C \\ 
     \midrule 
        0.0001 & 24.2  & 34.9  & 69.8  \\ 
        0.0005 & 22.7  & 33.6  & 68.0  \\ 
        0.001 & 22.4  & 33.5  & 66.7  \\ 
        0.005 & 27.1  & 38.0  & 70.6  \\ 
        0.01 & 30.7  & 41.0  & 78.8  \\ 
        0.05 & 38.1  & 48.8  & 94.1 \\ 
    \bottomrule[1pt]
    \end{tabular}
    \VspaceAfter
\end{table}

Lastly, we study the sensitivity of the learning rate in the range of [5$\times$10$^{-5}$, 0.0001, 0.0005, 0.001, 0.005, 0.01]. The model performance is stable when varying the learning rate. Nevertheless, too large learning rates lead to model performance collapse in the ImageNet-C task.
\begin{table}[!ht]
    \centering
        \caption{\textbf{Parameter analysis on the learning rate.} Average classification error (\%).
    }
    \label{table:paras_learning_rate}
    \small
    \VspaceBefore
    \renewcommand{\arraystretch}{0.8}
    \begin{tabular}{lccc}
    \toprule[1pt]
    learning rate & CIFAR10-C & CIFAR100-C & ImageNet-C \\ 
     \midrule 
        5$\times$10$^{-5}$ & 25.4  & 34.3  & 68.7  \\ 
        0.0001 & 25.3  & 34.0  & 68.3  \\ 
        0.0005 & 22.7  & 33.7  & 67.5  \\ 
        0.001 & 22.4  & 33.5  & 66.7  \\ 
        0.005 & 22.9  & 34.1  & 66.5  \\ 
        0.01 & 23.5  & 40.3  & 95.1 \\ 
    \bottomrule[1pt]
    \end{tabular}
    \VspaceAfter
\end{table}



\end{document}